\documentclass[]{bytedance_seed}
\microtypesetup{expansion=false}

\usepackage[toc,page,header]{appendix}
\usepackage{minitoc}
\usepackage{amsmath}
\usepackage{amssymb}

\graphicspath{{figs/}}

\newcommand{\TIF}{\textsc{Harness-IF}}

\title{Harness-IF: Evaluating Instruction Following Across Instruction Surfaces in Coding Agents}

\author[1,2,*]{Zining Huang}
\author[1,3,*]{Haoran Que}
\author[1,*]{Hong Zeng}
\author[1]{Ge Zhang}
\author[1]{Zuo Wang}
\author[1]{Jin Chen}
\author[1]{Haodong Wang}
\author[1,*]{Zhongfei Hou}
\author[1]{Changxin Pu}
\author[1,\dagger]{Shen Yan}
\author[1]{Wenhao Huang}

\affiliation[1]{ByteDance Seed}
\affiliation[2]{Tsinghua University}
\affiliation[3]{Peking University}

\contribution[*]{Work done at ByteDance Seed}
\contribution[\dagger]{Corresponding author}

\abstract{
When a coding agent obeys a rule, it may simply have been going to do that anyway. Existing instruction-following benchmarks cannot tell the difference: they concentrate rules in the user turn, while coding-agent benchmarks emphasize final task success. We introduce \textbf{Harness-IF}, which scores operational rules one at a time from execution evidence: $60$ realistic multi-turn coding items drawn from a $642$-rule library, $256$ rules receiving verdicts, placed on the five configurable surfaces a deployed agent reads. To separate compliance from coincidence we introduce Against-Prior Accuracy (\textbf{AP-Acc}), which scores only rules labeled as opposing unprompted defaults, observed by re-running tasks with the rule withheld across nine probe builds and curated otherwise. Across $12$ frontier models, accuracy spans $72.1$--$85.9\%$ and AP-Acc $66.1$--$78.6\%$; every model is worse on against-prior rules, by $3.6$ to $7.4$ points (mean $5.81$), and the direction survives a common-support analysis with item-clustered intervals. Aggregate scores therefore overstate compliance by a model-specific margin: prior control leaves the top build unchanged and exchanges three adjacent rank pairs. A counterbalanced conflict pilot on nine separate builds adds a second result: pooled precedence does not follow prompt depth, with system prompts, project files, and user instructions ahead of tool and skill descriptions.
}

\date{July 15, 2026}
\correspondence{Shen Yan at \email{sheny@bytedance.com}}

\begin{document}
\maketitle


\section{Introduction}
\label{sec:intro}

Coding agents operate under a stack of instructions while inspecting repositories, editing files, running commands, calling tools, and responding across multiple turns~\citep{google2026gemini31pro,singh2025openai,claudecode2024}. Compliance must therefore persist beyond a single response and across system prompts, tool and skill descriptions, project files such as \texttt{CLAUDE.md}, and user instructions. Existing instruction-following benchmarks concentrate rules in user prompts~\citep{zhou2023instructionfollowing,jiang2024followbench,qin2024infobench,wen2024complexbench}, whereas coding-agent benchmarks emphasize final task success~\citep{jimenez2024swebench,liu2023agentbench}. Neither directly reveals which operational rule an agent followed during a long harnessed workflow.

Instruction surface and instruction hierarchy are distinct. Hierarchy concerns which privileged source should prevail under conflict~\citep{wallace2024instructionhierarchy}; Harness-IF turns operational delivery surface into an evaluation dimension. Its controlled-relocation design holds rule meaning fixed while moving delivery across surfaces, enabling matched comparisons. We exercise this design in a controlled conflict pilot (E0) and complement it with a larger coding panel that assigns rules to operationally admissible surfaces, combining controlled identification with broad, realistic coverage.

\begin{figure}[!ht]
\centering
\includegraphics[width=\textwidth]{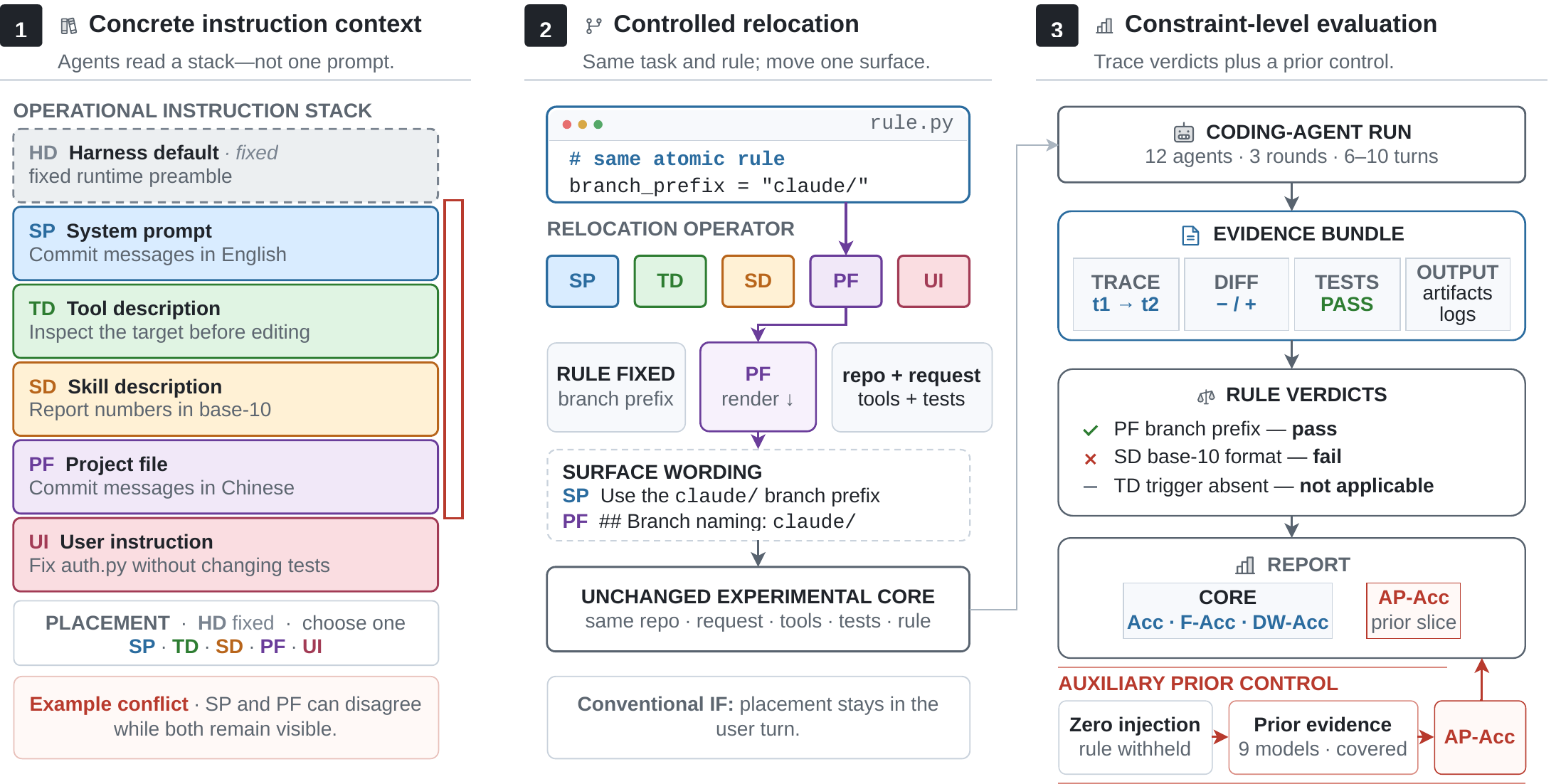}
\caption{Overview of Harness-IF. The left column shows an operational instruction stack. The middle column shows how the benchmark can relocate an atomic rule across configurable surfaces while holding its semantics fixed; the released main coding panel instead assigns suitable rules to surfaces. The right column records execution evidence, produces one verdict per applicable rule, and reports standard and prior-stratified metrics.}
\label{fig:concept}
\label{fig:concept-schema}
\end{figure}

We introduce \textbf{Harness-IF}, a benchmark that turns operational instruction following into a rule-level measurement problem: its $642$-rule library is instantiated as $60$ realistic multi-turn coding items scoring $256$ distinct rules, and every run yields a verdict per applicable rule rather than one task-level outcome. We then ask what those verdicts are worth. Against-Prior Accuracy (\textbf{AP-Acc}) scores only rules labeled as opposing unprompted defaults, and across $12$ frontier models it is lower than accuracy for every one. The inflation is not a constant that cancels when builds are compared: it varies twofold across the cohort. The resulting verdict traces also expose sharply different failure signatures across rule families and modalities.

The evaluation has three complementary components. The \emph{main coding panel} measures rule-level compliance and prior alignment across $12$ models. \emph{E0} isolates surface precedence under four counterbalanced conflicts and nine model builds. The \emph{non-coding extension} tests breadth on $40$ cases using a domain-appropriate case-macro metric.

Our contributions are:
\begin{itemize}
    \item \textbf{Benchmark that scores rules, not tasks:} a $642$-rule library of which $302$ rules are placed on the five configurable instruction surfaces of a deployed coding agent across $60$ realistic multi-turn items, and $256$ receive execution-grounded verdicts, one per applicable rule per run.
    \item \textbf{Metric that controls for unprompted defaults:} AP-Acc scores only rules labeled as opposing the unprompted default---observed in a zero-injection probe where that evidence is recoverable, and curated otherwise---separating instruction following from coincidence.
    \item \textbf{Evidence:} all $12$ models perform worse on against-prior rules under like-for-like and common-support analyses, while rule families and modalities exhibit distinct difficulty and failure signatures; E0 further reveals a robust pooled surface ordering that is inconsistent with simple prompt-depth accounts.
\end{itemize}


\section{Related Work}
\label{sec:related}

\paragraph{Instruction-Following Evaluation.}
Instruction-following (IF) evaluation has moved from coarse preference
judgment to explicit constraint checking. IFEval
\citep{zhou2023instructionfollowing} made verifiable constraints a
standard protocol; FollowBench \citep{jiang2024followbench}, InfoBench
\citep{qin2024infobench}, and ComplexBench
\citep{wen2024complexbench} expanded the space to graded difficulty,
information constraints, and multi-constraint composition. The most
recent benchmarks test harder prompts rather than only more prompts:
Multi-IF \citep{he2024multiif} adds multilingual multi-turn dialogue,
CFBench \citep{liu2025cfbench}, LIFBench \citep{shi2025lifbench}, and
EIFBench \citep{gupta2025eifbench} stress complex or long-context
constraint sets, IFBench \citep{pyatkin2025ifbench} broadens
verifiable rule checking, and AgentIF \citep{qi2025agentif} introduces
agentic IF scenarios. These benchmarks establish constraint-level
measurement, but their experimental variable is still mainly the
instruction content or scenario. Harness-IF asks a different question:
when the same rule is delivered through different harness surfaces, does
the agent still follow it?

\begin{table}[!ht]
\centering
\caption{Comparison with representative recent benchmarks. Surfaces is the
number of distinct instruction-delivery surfaces a benchmark represents;
for Harness-IF this is six (HD, SP, TD, SD, PF, UI), of which the harness
default (HD) is fixed and the remaining five are configurable placement
surfaces: placement is varied as an experimental variable in the E0
conflict pilot and assigned by admissibility in the main panel. Prior
control means explicit control for unprompted default behavior.
\checkmark{} marks full support, $\bullet$ partial support, and --
absence.}
\label{tab:compare}
\small
\begin{tabular}{@{}l *{5}{c}@{}}
\toprule
\multirow{2}{*}{Benchmark} & \multirow{2}{*}{\textbf{Surfaces}} &
\multirow{2}{*}{\textbf{Multi-Turn}} & \textbf{Tool} &
\textbf{Prior} & \textbf{Rule-Level} \\
& & & \textbf{Use} & \textbf{Control} & \textbf{Scoring} \\
\midrule
IFEval \citep{zhou2023instructionfollowing}       & 1 & --         & --         & --         & \checkmark \\
ComplexBench \citep{wen2024complexbench}          & 1 & --         & --         & --         & \checkmark \\
IFBench \citep{pyatkin2025ifbench}                & 1 & $\bullet$  & $\bullet$  & --         & \checkmark \\
AgentIF \citep{qi2025agentif}                     & 2 & $\bullet$  & $\bullet$  & --         & \checkmark \\
CodeIF-Bench \citep{wang2025codeifbench}          & 1 & \checkmark & $\bullet$  & --         & \checkmark \\
BFCL \citep{patil2025bfcl}                        & 1 & \checkmark & \checkmark & --         & --         \\
AppWorld \citep{trivedi2024appworld}              & 1 & \checkmark & \checkmark & --         & $\bullet$  \\
$\tau^2$-bench \citep{barres2025tau2}             & 2 & \checkmark & \checkmark & --         & $\bullet$  \\
SWE-Bench Pro \citep{deng2025swebenchpro}         & 1 & \checkmark & \checkmark & $\bullet$  & --         \\
Terminal-Bench \citep{merrill2025terminalbench}   & 1 & \checkmark & \checkmark & $\bullet$  & --         \\
\midrule
\textbf{Harness-IF} & \textbf{6} & \checkmark & \checkmark & \checkmark & \checkmark \\
\bottomrule
\end{tabular}
\end{table}

\paragraph{Agentic and Coding Evaluation.}
Agent benchmarks now cover realistic environments in which models must
plan, call tools, inspect state, and recover from intermediate errors.
AgentBench \citep{liu2023agentbench} and GAIA \citep{mialon2024gaia}
evaluate general-purpose assistants; WebArena \citep{zhou2024webarena},
OSWorld \citep{xie2024osworld}, BFCL \citep{patil2025bfcl},
AppWorld \citep{trivedi2024appworld}, $\tau$-bench
\citep{yao2024taubench}, and $\tau^2$-bench \citep{barres2025tau2}
evaluate web, desktop, tool-calling, application, and user-interaction
agents. A parallel line evaluates technical work---repository repair, command-line work, and ML engineering---through the SWE-bench family \citep{jimenez2024swebench,openai2024swebenchverified,yang2024sweagent,yang2024swebenchmm,deng2025swebenchpro}, Terminal-Bench \citep{merrill2025terminalbench}, and the MLAgentBench/MLE-bench/PaperBench line \citep{huang2024mlagentbench,chan2024mlebench,openai2025paperbench}. These benchmarks are high fidelity, but most headline scores collapse a trajectory into task success. Harness-IF is complementary: it uses realistic coding-agent execution, but treats instruction following itself as the object of measurement.

\paragraph{Harness and Trajectory Evaluation.}
For deployed agents, the model is only one part of the system: behavior
is shaped by system prompts, tool schemas, skills, project files,
memory, retries, and the trace visible to the evaluator. Toolformer
\citep{schick2023toolformer} and Gorilla \citep{patil2024gorilla} show
why tool context matters; instruction-hierarchy work
\citep{wallace2024instructionhierarchy} and IHEval's synthetic conflicts
over system, user, history, and tool-output messages
\citep{zhang2025iheval} show why authority context does. IHEval tests compliance with a prescribed message hierarchy; Harness-IF instead measures distributed constraints in executable workspaces, and E0 estimates precedence over a broader surface set without assuming a universal order. New agent-evaluation papers
make the same point from different angles: MCP tool descriptions affect
agent efficiency \citep{hasan2026mcpdescriptions}, open skill manifests
require auditing \citep{ying2026openskilleval}, process defects can
survive final-task evaluation \citep{he2026procbench}, and long-horizon
coding agents can exploit benchmark specifications
\citep{zhao2026specbench}. Harness-IF turns that conclusion into a
surface-aware benchmark design: it records rule-level verdicts and
reports AP-Acc to separate prompted compliance from prior-aligned
default behavior.


\section{Harness-IF}
\label{sec:design}

In a complex agentic workflow, the instruction is not a single standalone user request. As shown in Figure~\ref{fig:concept-schema}, the agent reads multiple prompts in sequence; we refer to these prompt levels as instruction surfaces. Requirements on different surfaces can conflict, yet how agents follow the full instruction stack remains under-explored~\citep{qi2025agentif,wen2024complexbench}. We therefore build Harness-IF around where a rule is delivered; Table~\ref{tab:compare} summarizes how it differs from prior benchmarks.

\begin{figure}[!ht]
\centering
\includegraphics[width=\textwidth]{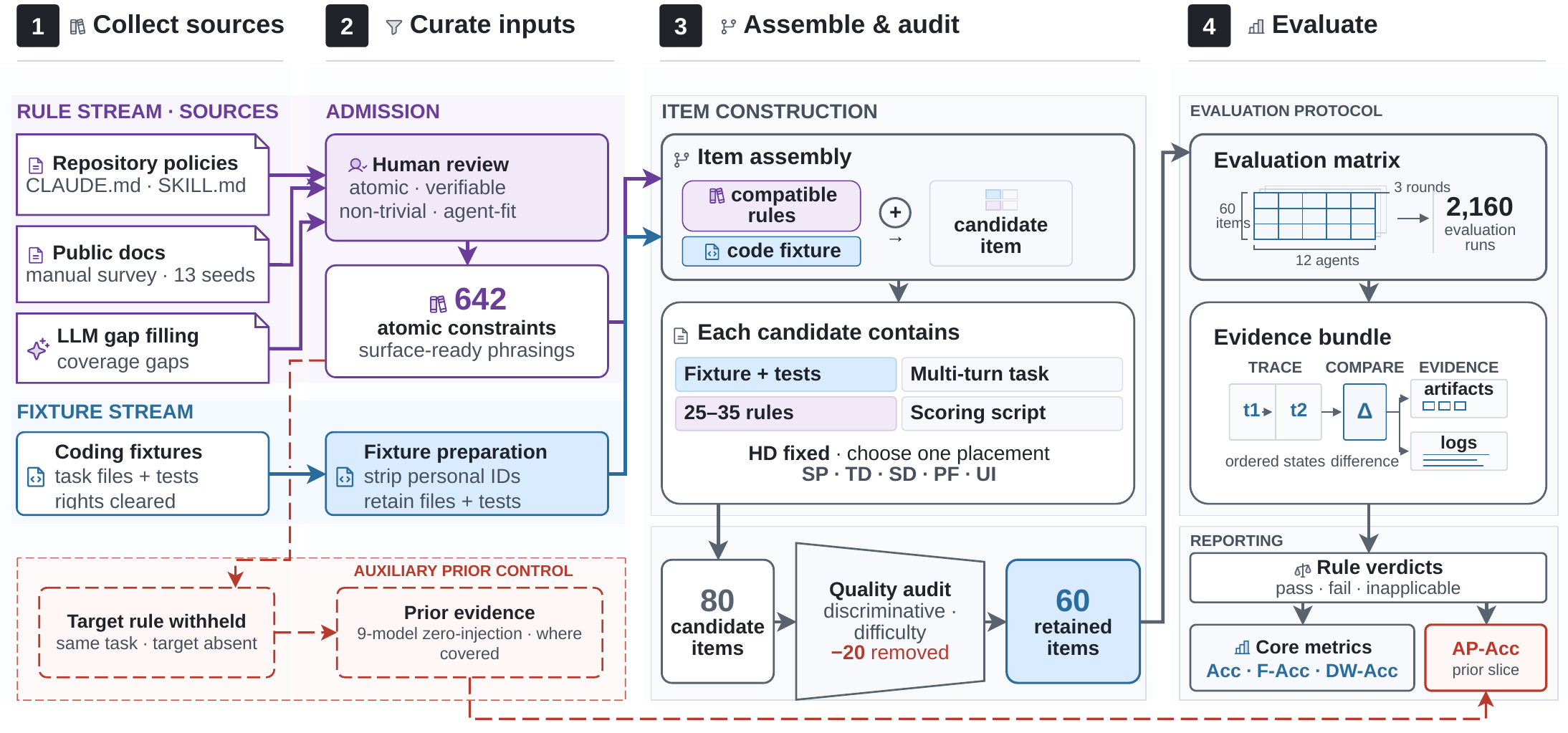}
\caption{The Harness-IF construction and evaluation pipeline. Candidate constraints and coding fixtures are curated and assembled into 60 quality-audited multi-turn items. The resulting $60\times12\times3=2{,}160$ agent--item--round runs produce traces, state differences, artifacts, and logs for rule-level verdicts and core/AP-Acc reporting. The auxiliary zero-injection branch supplies prior evidence without entering the core metrics.}
\label{fig:construction}
\end{figure}

\subsection{Instruction Surfaces}
\label{sec:design:surfaces}

Following the common configuration used by modern coding agents~\citep{claudecode2024,anthropic2024effectiveagents}, we distinguish six instruction surfaces:

\begin{itemize}
\item \textbf{Harness Default (HD).} The default instructions inserted by the agent platform at the start of each run; users usually cannot edit them in deployment.
\item \textbf{System Prompt (SP).} Instructions written by the system developer to set the agent's role and general behavior, such as ``never expose secrets'' or ``answer with high confidence.''
\item \textbf{Tool Description (TD).} Instructions that describe what a tool does and how the agent should use it.
\item \textbf{Skill Description (SD).} Instructions that describe when a reusable skill should be used and what rules the agent should follow when using it.
\item \textbf{Project File (PF).} Project-level instructions stored in files such as \texttt{CLAUDE.md}, \texttt{CONTRIBUTING.md}, or \texttt{AGENTS.md}.
\item \textbf{User Instruction (UI).} The user's current request to the agent.
\end{itemize}

These surfaces are written by different parties, such as platform developers, tool authors, project maintainers, and users, and appear in different parts of the prompt. 
Formally, let $x$ be a single constraint, $y$ be the agent output, and $s$ be the surface on which the constraint is placed. We write $F(x,y,s)\in\{\textsf{pass},\textsf{fail},\textsf{n/a}\}$ for the rule-level evaluator that judges whether output $y$ satisfies constraint $x$ when $x$ is delivered on surface $s$; \S\ref{sec:design:accuracy} turns its pass/fail outcomes into the indicator $z$ used by the metrics.
Most benchmarks~\citep{wen2024complexbench,deng2025swebenchpro} keep $s$ fixed at the user instruction.
Harness-IF evaluates how well agents follow specific constraints placed on five configurable surfaces, $s\in\{\textsc{SP},\textsc{TD},\textsc{SD},\textsc{PF},\textsc{UI}\}$.

\subsection{Data Structure}
\label{sec:design:schema}

\paragraph{Constraints.}
A constraint is one specific rule that an agent is asked to follow. Each rule is narrow enough to be judged pass, fail, or not applicable from the execution evidence of a run. The seven families are professional writing, output control, code style, workflow, quantitative limits, conditional logic, and tool use.

\paragraph{Surfaces.}
A constraint is admissible on a surface when that surface's authoring role could plausibly carry the rule in deployment: a branch-naming rule can sit in a project file or a system prompt, but not in a tool schema. When several surfaces are admissible we sample one uniformly, using surface-appropriate phrasings that preserve rule semantics. This design supports descriptive surface stratification in the main panel; E0 provides the controlled conflict comparison.

\paragraph{Scenarios.}
Each scenario provides working files for a realistic coding task---fixing a backend API bug, updating a frontend component, editing a data-processing script, writing a test, or changing project documentation---so items differ in language, file layout, and the kind of edit required. The evaluated panel draws on eight scenarios from a $13$-scenario library, spanning backend, frontend, systems, data/ML, automation, security testing, tool orchestration, and technical documentation.

\begin{figure}[!ht]
\centering
\includegraphics[width=0.94\linewidth]{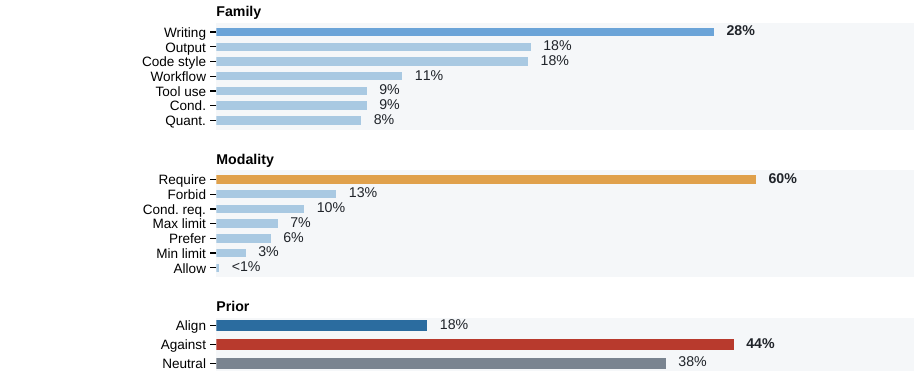}
\caption{Composition of the 642-rule library across three independent annotation axes: rule family, logical modality, and behavioral prior. Prior labels indicate whether a rule is aligned with, against, or neutral to observed or curated unprompted behavior; Cond. req. denotes conditional requirement. The figure describes the library, not the evaluated subset: the coding panel instantiates $302$ of these rules and scores $256$ of them, and professional-writing rules are exercised only in the non-coding extension.}
\label{fig:constraint-dist}
\end{figure}

Each data item contains a code scenario, multi-turn user instructions, selected constraints, and their assigned surfaces.
Figure~\ref{fig:constraint-dist} shows the library's composition. Requirements are the dominant modality ($60\%$), and professional writing is the largest single family ($28\%$) even though it is exercised outside the coding panel. Prior labels are deliberately mixed---$18\%$ align-prior, $44\%$ against-prior, and $38\%$ neutral---so the panel cannot be passed by defaults alone. Scenario is a separate annotation axis and is not counted in this figure.

\subsection{Data Curation}
\label{sec:design:annotation}

As shown in Figure~\ref{fig:construction}, the data curation process has four steps: we collect constraints from raw sources, filter them, format them into benchmark items, and apply quality checks.

\paragraph{Raw Sources.}
\begin{sloppypar}
The constraint library began with a manual survey of publicly accessible GitHub software projects, focusing on project-instruction and contributor documents such as \texttt{CLAUDE.md},\allowbreak{} \texttt{AGENTS.md},\allowbreak{} \texttt{CONTRIBUTING.md},\allowbreak{} skill descriptions, and tool schemas across multiple programming languages and software-engineering workflows. We atomized recurring requirements into benchmark-ready rules, normalized them across surface-specific phrasings, and used reviewed LLM-assisted proposals plus author review to fill under-covered taxonomy cells. The resulting tasks, specifications, and scoring artifacts are project-authored rather than direct copies of source repositories. Detailed provenance is maintained for internal audit and rights review, but the public artifact exposes only coarse source categories so individual constraints cannot be joined directly to named repositories. Coding workspaces were assembled to match realistic technology stacks and are included in the release following the 2026-07-27 owner attestation. Appendix~\ref{sec:datasheet} provides the detailed release boundary.
\end{sloppypar}

\paragraph{Filtering and Item Formatting.}
Candidate constraints must be atomic, verifiable, non-trivial, and suitable for coding-agent tasks; proposals are deduplicated and human-reviewed before admission, with targeted expansion for under-covered categories. For each scenario and set of working files we then select constraints and place each one on a single admissible surface.

\paragraph{Quality Filtering.}
From $80$ candidates, quality review retained $60$ items. Because selection used observed difficulty and discriminativeness, the reported panel may favor items that separated the pilot models (selection optimism); Appendix~\ref{sec:datasheet} documents the full disposition and validation sequence.

In total, the Harness-IF library contains $642$ atomic constraints and the evaluated panel comprises $60$ multi-turn items. Each item injects a pack of $25$--$35$ rules, of which $10$--$27$ are scorable given the opportunities the item creates; across the panel, $302$ distinct library rules are instantiated and $256$ receive at least one verdict.

\subsection{AP-Acc}
\label{sec:design:accuracy}

We use rule-level accuracy metrics to measure whether agents follow the selected constraints. For each agent output, the evaluator judges each constraint as passed, failed, or not applicable. Let $z_{a,i,r}=1$ if agent $a$ satisfies constraint $r$ in item $i$, and $z_{a,i,r}=0$ otherwise. Let $\mathcal{E}_a$ be the set of pass/fail constraint instances for agent $a$; not-applicable constraints are excluded from the denominator because the item does not create a valid opportunity to judge whether the agent followed that constraint.

For a like-for-like binary recomputation, the basic accuracy treats all eligible constraint instances equally:
\begin{align}
\mathrm{Acc}(a)
&=
\frac{\sum_{(i,r)\in\mathcal{E}_a} z_{a,i,r}}
     {|\mathcal{E}_a|}.
\end{align}

We also report two cohort-adaptive diagnostics. First, filtered accuracy (\textbf{F-Acc}) removes item--rule pairs that do not distinguish agents in the evaluated cohort. Let $\mathcal{D}$ be the set of item--rule pairs that produce different pass/fail outcomes across agents:
\begin{align}
\mathrm{F\mbox{-}Acc}(a)
&=
\frac{\sum_{(i,r)\in\mathcal{E}_a\cap\mathcal{D}} z_{a,i,r}}
     {|\mathcal{E}_a\cap\mathcal{D}|}.
\end{align}
Second, discrimination-weighted accuracy (\textbf{DW-Acc}) gives more weight to rules that better separate the evaluated agents. Let $d_r\geq 0$ be the non-negative Pearson correlation between agents' pass rates on rule $r$ and their overall accuracy. Rules with non-positive or undefined discrimination receive $d_r=0$:
\begin{align}
\mathrm{DW\mbox{-}Acc}(a)
&=
\frac{\sum_{(i,r)\in\mathcal{E}_a} d_r z_{a,i,r}}
     {\sum_{(i,r)\in\mathcal{E}_a} d_r}.
\end{align}

Accuracy and AP-Acc can differ when a rule matches default behavior. We therefore assign behavioral prior labels using zero-injection evidence and curated prior annotations. AP-Acc is a behavioral stratification rather than a training-provenance claim; Appendix~\ref{sec:app:scoring} reports label lineage and sensitivity analyses (Figure~\ref{fig:constraint-dist}).

We introduce against-prior accuracy (\textbf{AP-Acc}), which only evaluates constraints in the against-prior set $\mathcal{P}$. This metric reports performance on constraints labeled as opposing the corresponding default behavior:
\begin{align}
\mathrm{AP\mbox{-}Acc}(a)
&=
\frac{\sum_{(i,r)\in\mathcal{E}_a,\ r\in\mathcal{P}} z_{a,i,r}}
     {|\{(i,r)\in\mathcal{E}_a: r\in\mathcal{P}\}|}.
\end{align}

All four displayed columns are computed from the released $2{,}160$-record panel under the equations above, over the same eligible verdicts, so $\Delta$ is a like-for-like contrast rather than a difference between aggregation conventions. F-Acc and DW-Acc are cohort-relative diagnostics and are interpreted descriptively: both change if the evaluated cohort changes. Exact denominators, the common-support analysis, and the per-rule discrimination weights appear in Appendix~\ref{sec:app:scoring}, and the release includes the script that regenerates every displayed number from the shipped verdict records.



\section{Results}
\label{sec:results}

\subsection{Main Benchmark Results}
\label{sec:results:leaderboard}

\begin{table}[!ht]
\centering
\caption{Rule-level accuracy (\%) across 12 model builds, recomputed from the released $2{,}160$-record panel. All four columns are binary rates over the same eligible verdicts, so $\Delta=\mathrm{Acc}-\mathrm{AP\mbox{-}Acc}$ is a like-for-like contrast. Rows are ordered by unrounded Acc, which separates MiniMax-M2.7 from Seed-2.0-Pro by $0.03$ points; bold marks each column's best value.}
\label{tab:leaderboard}
\small
\begin{tabular}{@{}l c c c c c@{}}
\toprule
\textbf{Models} & \textbf{Acc} & \textbf{F-Acc} & \textbf{DW-Acc} & \textbf{AP-Acc} & \textbf{$\Delta$} \\
\midrule
Claude-Opus-4.7~\citep{anthropic2026modeldocs} & \textbf{85.9} & \textbf{79.3} & \textbf{88.5} & \textbf{78.6} & $+7.3$ \\
GPT-5.5~\citep{openai2026gpt55}   & 83.1 & 75.5 & 81.2 & 77.0 & $+6.1$ \\
Claude-Sonnet-4.6~\citep{anthropic2026modeldocs} & 82.5 & 73.9 & 82.2 & 78.5 & $+4.0$ \\
Claude-Haiku-4.5~\citep{anthropic2026modeldocs}  & 79.0 & 69.4 & 75.9 & 71.9 & $+7.2$ \\
GLM-5.1~\citep{zai2026glm51}           & 78.8 & 67.9 & 75.6 & 75.2 & $+3.6$ \\
Qwen-3.6-Max~\citep{qwen2026qwen36max}      & 76.7 & 65.9 & 70.5 & 71.7 & $+5.1$ \\
Hy3~\citep{tencent2026hy3}         & 76.2 & 65.2 & 69.7 & 70.8 & $+5.4$ \\
Kimi-K2.6~\citep{moonshot2026kimi26}         & 76.1 & 65.0 & 70.3 & 70.0 & $+6.1$ \\
Gemini-3.1-Pro~\citep{google2026gemini31pro}    & 75.4 & 64.0 & 70.0 & 69.8 & $+5.6$ \\
MiniMax-M2.7~\citep{minimax2026m27}      & 73.6 & 61.2 & 65.1 & 67.7 & $+5.9$ \\
Seed-2.0-Pro~\citep{bytedance2026seed2}      & 73.6 & 61.1 & 63.2 & 66.1 & $+7.4$ \\
StepFun-3.5~\citep{stepfun2026step35flash}       & 72.1 & 59.1 & 62.5 & 66.2 & $+5.9$ \\
\bottomrule
\end{tabular}
\end{table}

\paragraph{Evaluation scale.}
The main coding panel evaluates $12$ frontier model builds on $60$ items over three rounds: $2{,}160$ agent--item--round runs and $40{,}104$ rule-level verdict rows, one per applicable constraint per run. Deterministic checks (regex, AST, cross-file, command-output) cover $13.3\%$ of eligible verdicts; a GPT-5.2 judge~\citep{singh2025openai} scores rubric constraints by three-vote majority and adjudicates hybrid checks, so $86.8\%$ of rows involve the judge. The like-for-like binary analysis contains $37{,}616$ eligible verdicts, including $19{,}449$ against-prior verdicts. Appendix~\ref{sec:app:expdetail} tables the exact build identifier behind each row and documents the serving configuration, run status, and exclusion rules.

Table~\ref{tab:leaderboard} supports two conclusions. First, every model is less successful on the against-prior subset, so aggregate scores overstate compliance where a rule departs from model defaults. Second, the overstatement is model-specific: it ranges from $3.6$ to $7.4$ points, a twofold spread. Claude-Opus-4.7 leads all four columns, so prior control does not change the top-ranked build, but it exchanges three adjacent rank pairs (2--3, 4--5, and 11--12). Accuracy spans $13.7$ points across the cohort, with a standard deviation of $4.2$ points.

Models largely agree on which rules are hard: correlating each model's per-rule pass-rate vector against the cohort mean, over the $242$ rules every model attempted, gives $0.57$--$0.89$ (mean $0.80$). The panel therefore measures a shared difficulty ordering, which is what makes the adjacent-rank exchanges informative rather than noise.

\paragraph{Against-prior rules expose a consistent compliance gap.}
Every evaluated model scores lower on the against-prior subset. Under the like-for-like binary definition, the mean Acc--AP-Acc gap is $5.81$ points across $37{,}616$ eligible verdicts and remains positive for all $12$ models. A stricter common-support analysis retains $2{,}430$ of $3{,}342$ (item, round, rule) observation keys ($72.7\%$) on which every model produced a clean pass/fail outcome; item-clustered $95\%$ intervals for the paired gap remain positive model by model. On this fixed panel the direction is stable: for all $12$ builds, aggregate instruction-following scores exceed performance on rules that oppose observed defaults.

\paragraph{The benchmark resolves broad patterns more clearly than adjacent ranks.}
On common support, item-clustered intervals leave every adjacent model comparison unresolved, so we treat the displayed order as a point ranking while resting the paper's claims on the prior-alignment and failure-pattern results.



\subsection{Failures Concentrate on Rules That Demand Action}
\label{sec:results:errors}

\begin{figure}[!ht]
\centering
\includegraphics[width=0.72\linewidth]{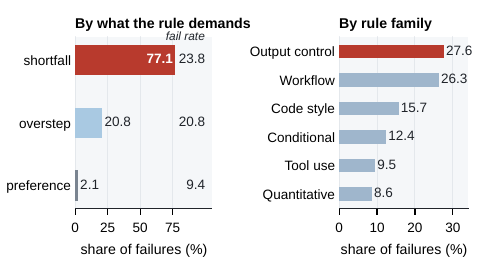}
\caption{Failure decomposition over the released panel. Left: share of the $8{,}440$ failures by what the violated rule demands, beside each class's own failure rate; shortfall rules carry most of the mass at a comparable rate, so the gap is one of exposure. Right: share of failures by rule family.}
\label{fig:error-modes}
\end{figure}

We group failures by what the violated rule demands, which is recoverable from the released records without free-text classification: a rule that requires an action or sets a minimum can only fail by the agent falling short, and a rule that forbids an action or caps a quantity can only fail by the agent overstepping. Shortfall rules absorb $77.1\%$ of the $8{,}440$ failures, against $20.8\%$ for overstep rules and $2.1\%$ for soft preferences (Figure~\ref{fig:error-modes}). The asymmetry reflects exposure rather than propensity: the panel contains $27{,}306$ shortfall instances against $8{,}443$ overstep instances, and the two classes fail at similar rates ($23.8\%$ versus $20.8\%$). Most compliance failures in realistic coding work are therefore omissions of demanded behavior, so verifiers tuned to detect excess output address only about a fifth of the failure mass. Appendix~\ref{sec:app:errors} gives the full decomposition.

Failure mass is unevenly distributed across families. Output control and workflow together account for $53.9\%$ of all failures ($27.6\%$ and $26.3\%$), followed by code style ($15.7\%$), conditional logic ($12.4\%$), tool use ($9.5\%$), and quantitative limits ($8.6\%$). Output control combines the largest failure mass with the lowest pass rate, which makes it the most productive remediation target; workflow's mass instead follows from its size in the panel, since its pass rate is mid-range.

\subsection{Commands and Output-Control Rules Are Hardest}
\label{sec:results:difficulty}

\begin{figure}[!ht]
\centering
\includegraphics[width=0.72\linewidth]{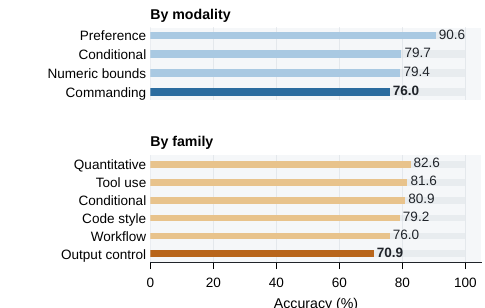}
\caption{Pooled binary accuracy over the $12$ builds by modality (top) and by the six families that receive verdicts (bottom); the lowest bar in each panel is emphasized. Commanding constraints are hardest by modality, and output control is the lowest-scoring family overall and for $11$ of the $12$ builds.}
\label{fig:difficulty}
\end{figure}

Difficulty varies substantially across both logical modality and rule family (Figure~\ref{fig:difficulty}); Appendix~\ref{sec:app:axes} defines these axes. For difficulty reporting we pool the seven governed modality operators into four coarse classes: Commanding (require/forbid), Conditional (conditional-require), Quantitative (limit-max/limit-min), and Preference (prefer/allow); the coarse ``Quantitative'' class is the pooled modality and is distinct from the like-named rule family. The released coding scorecard provides family aggregates for six of the seven library families; professional writing is not included in this comparison.

Commanding rules are hardest by modality ($76.0\%$), against $79.4\%$ for numeric bounds, $79.7\%$ for conditional rules, and $90.6\%$ for preferences. The preference result is consistent with default alignment: preferences can match existing behavior, whereas commands often require departure from it. Family differences are narrower: output control scores $70.9\%$ against $82.6\%$ for quantitative limits, an $11.7$-point spread; it is the lowest-scoring family for $11$ of the $12$ builds, and no build clears $79\%$ on it.

\subsection{Surface Precedence Does Not Follow Prompt Depth}
\label{sec:results:channels}

\begin{figure}[t]
\centering
\includegraphics[width=0.40\linewidth]{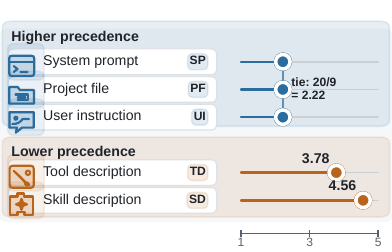}
\caption{E0 mean conflict ranks over nine older model builds (lower = higher precedence). SP, PF, and UI are an exact three-way tie: each model-level rank sum is $20$, so each mean is $20/9=2.22\ldots$; TD $=3.78$ and SD $=4.56$. This deterministic-only pilot is separate from the main 12-model coding panel; ranks summarize four synthetic conflict pairs and are not pass rates.}
\label{fig:surface-ranks}
\end{figure}

E0 provides a controlled, scoped test of surface precedence. System prompts, project files, and user instructions tie exactly for the best mean rank: each has a model-level rank sum of $20$, hence $20/9=2.22\ldots$. They precede tool descriptions at $3.78$ and skill descriptions at $4.56$ (Figure~\ref{fig:surface-ranks}). By prompt depth we mean position in the assembled context, with the user turn last; a depth account predicts that later-placed instructions win. That the user instruction only ties for first is inconsistent with such an account.

E0 contains $916$ runs from nine older model builds on four synthetic conflict pairs, uses deterministic-only scoring, and is not pooled with the main coding panel. Counterbalancing assigns $458$ runs to each direction; $889$ produce decisive outcomes. A pooled Bradley--Terry analysis places SP, PF, and UI above TD, with SD last. The ordering survives separate direction fits, equal-cell weighting, all leave-one-pair/model-out fits, and four conservative assignments of the $27$ errors. A crossed model-and-pair bootstrap preserves the complete ordering in $9{,}652/10{,}000$ resamples; only $6/9$ individual-build fits reproduce it exactly, so we interpret it as a pooled cross-build tendency, not a universal hierarchy (Appendix~\ref{sec:app:exp:e0}). The main panel instead assigns suitable rules to surfaces and supports descriptive stratification rather than a paired surface effect.


\subsection{Reliability: Cross-Model Patterns Are Stable, Rank Margins Are Not}
\label{sec:results:reliability}

The prior-alignment gap stays positive for all $12$ models in the fully deterministic subset, averaging $13.09$ points over $5{,}013$ eligible verdicts; it is larger there because that subset over-represents the pattern- and command-checked families, against $5.81$ points panel-wide. Test--retest ICC is $0.725$ across models and $0.599$ across agent--item cells, and an earlier human-reference audit, run under a five-vote rather than the released three-vote configuration, showed $69.0\%$ agreement ($\kappa=0.515$ over $919$ rows). Verdicts are far less stable under a judge swap ($62.1\%$ agreement, $\kappa=0.163$ on $116$ paired clean verdicts). Since $86.8\%$ of verdict rows involve the judge, this is the dominant uncertainty in the measurement, and we therefore report cross-model patterns and treat the displayed order as a point ranking. Appendix~\ref{sec:app:reliability} provides the complete calibration protocols. On $n=65$ common non-error samples, Claude--GPT inter-LLM agreement is $\kappa=0.4717$, which is not human validation.

\paragraph{The instrument transfers beyond code, the ranking does not.}
A separate exploratory panel scores $40$ non-coding cases across five domains over $1{,}428$ valid trajectories from the same $12$ builds. Case-macro pass rates span $65.8$--$84.8\%$, led by GPT-5.5 rather than the coding leader; that panel's metric and population differ and are never pooled with the coding results (Appendix~\ref{sec:app:noncoding}).




\section{Limitations}
\label{sec:limitations}

\paragraph{Scope.}
Harness-IF targets multi-turn coding agents. The $60$ evaluated items were selected from an $80$-item working set through quality and discriminativeness review, and they instantiate $302$ of the library's $642$ rules and score $256$ of them; professional-writing rules are exercised only in the non-coding extension. Our claims therefore hold for this panel of items and models; the $40$-case non-coding extension provides complementary breadth under a separate case-macro metric.

\paragraph{Measurement.}
AP-Acc is a behavioral stratification over observed or curated defaults, not a claim about model training provenance. We report label lineage in the appendix, including the overlap between the zero-injection probe cohort and the evaluated panel, and use one like-for-like binary definition throughout. Because $86.8\%$ of verdicts involve an LLM judge whose labels shift under a judge swap, absolute levels are instrument-specific and the cross-model comparisons, which share the instrument, carry the claims. F-Acc and DW-Acc are cohort-adaptive diagnostics, while common-support and item-clustered analyses provide the primary denominator and uncertainty checks. The failure decomposition groups violations by what the rule demands and therefore describes rule structure rather than latent causal mechanisms.

\paragraph{Resolution and surfaces.}
Calibration and test--retest analyses support the benchmark's cross-model patterns but not fine distinctions among neighboring models. The main panel places rules on operationally admissible surfaces for realistic coverage; E0 separately supplies the controlled comparison. Its pooled ordering is robust across crossed-bootstrap, direction, deletion, weighting, and error analyses and is interpreted as an E0 cross-build tendency rather than a universal hierarchy.


\section{Conclusion}
\label{sec:conclusion}

Harness-IF makes operational instruction following measurable at the level of individual rules and delivery surfaces. Across $60$ multi-turn coding items, $256$ scored rules, and $12$ frontier models, every model performs worse on against-prior rules under both full-panel and common-support analyses, revealing compliance differences that aggregate scores obscure. The benchmark further localizes failure signatures by family and modality, while its counterbalanced E0 study identifies a robust pooled ordering in which system prompts, project files, and user instructions lead tool and skill descriptions. Harness-IF therefore provides a unified measurement framework for determining which operational rules an agent follows, where those rules are delivered, and how reliably they survive execution in realistic workflows. Coding-agent evaluation should treat instruction compliance as an execution-level object: preserve rule provenance, score individual opportunities, separate against-prior behavior from aggregate success, and counterbalance delivery surfaces when making precedence claims.

\section*{Data and Code Availability}
\label{sec:availability}

The benchmark is prepared for public release as a self-contained package: the $642$-rule constraint library in YAML, the $60$ assembled coding items with their scenario fixtures and ground-truth scoring scripts, the $2{,}160$-record verdict panel behind every number reported here, and the evaluation and analysis code. A single script recomputes each displayed figure and table from the shipped verdict records, so the results in this paper can be reproduced offline, without model API access and without re-running any agent. Evaluating a new model additionally requires a coding-agent harness and API access to both the model under test and a judge model.

Project-authored software is released under Apache-2.0, and the project-authored benchmark text, data, and sanitized derived results under CC-BY-4.0. Raw provider outputs and agent run workspaces are not redistributed. The release location will be given in a later version of this preprint.

\clearpage
\bibliographystyle{plainnat}
\bibliography{references,references_nlp}

\clearpage
\beginappendix

\section{Design axis definitions}
\label{sec:app:axes}

Each of the 642 atomic constraints in the \TIF{} library is tagged along eight separable axes.

\begin{description}
\item[Family (7).] The rule-family annotation: professional-writing, output-control, code-style, workflow, quantitative, conditional-logic, and tool-use. Family is independent of scenario applicability. Synthetic examples include requiring a verification step after implementation (workflow) and requiring structured rather than free-form tool parameters (tool-use).

\item[Modality (7).] The logical operator: require, forbid, conditional-require, limit-max, limit-min, prefer, allow. Modality captures difficulty structure that family alone misses: a forbid and a require on the same content are different instructions. Synthetic examples include requiring a short completion summary, forbidding generated cache files, requiring validation when a configuration changes, and preferring a compact report format.

\item[Prior (3).] The model's observed or curated default behavior in the absence of the instruction: align-prior (the instruction matches the default tendency), against-prior (the instruction pushes against it), neutral. A zero-injection ablation runs each task with the target rule withheld across nine probe builds; a rule receives a consensus label when at least five of those nine agree, which is recoverable for 287 rules. Other final labels use recoverable pre-existing curation or have unknown lineage, as detailed in the full reliability analysis. This probe cohort is the same set of nine builds used in the E0 conflict pilot, run there on different tasks for a different purpose; the two experiments are separate and their results are never pooled. Five of the nine share an identifier with a model in the evaluated panel, and the full reliability analysis quantifies that overlap. The current governed totals are 115 align-prior, 282 against-prior, and 245 neutral. These labels are behavioral strata, not causal claims about training provenance.

\item[Observability (4).] Where compliance is visible: surface (tokens in the final turn), structural (file layout, AST), behavioral (test pass/fail, side effects), deep (semantic intent). Synthetic examples include a required summary heading (surface), a required module layout (structural), successful local validation (behavioral), and preservation of an argument's causal order (deep).

\item[Verifiability (3).] How compliance can be checked: deterministic (regex, AST match, cross-file equality), rubric (LLM judge over a written rubric), subjective (human-only, currently unused in scoring). Synthetic examples include regex over a generated heading (deterministic), a pass-if/fail-if rubric for concise explanatory prose (rubric), and aesthetic fluency (subjective, not scored).

\item[Universality (4).] Applicability: universal, cross-coding, cross-non-coding, specific. This axis governs which scenarios a constraint can be composed into: a completion summary can be universal, local validation can be cross-coding, an audience-specific call to action can be cross-non-coding, and a framework toggle can be scenario-specific.

\item[Surface fit.] Each constraint carries a per-surface suitability score \{none, low, medium, high\} indicating whether it can be placed in each surface, preventing semantically impossible placements. For example, a synthetic tool-parameter rule fits naturally in TD or PF, may be stated globally in SP, and is less natural as an ad hoc UI request.

\item[Surface variants.] Each applicable surface has a pre-authored rendering that preserves semantics while matching that surface's role. For a synthetic compact-summary rule, SP may state ``Keep generated summaries compact,'' PF may state ``Output style: compact summaries,'' and UI may request ``Return a compact completion summary.''
\end{description}

The (family $\times$ modality $\times$ prior $\times$ observability $\times$ verifiability $\times$ universality) combinatorial surface is $7 \times 7 \times 3 \times 4 \times 3 \times 4 = 7056$ cells. Our 642 constraints populate roughly $420$ of these cells after deduplication on rhetorical content; the remaining cells are either semantically empty (e.g., forbid $\times$ subjective) or under-populated in the current library.

\subsection*{Authoring waves}

A hash-pinned historical artifact verifies that public project-instruction and contributor documents informed the initial authoring wave, and a later commit preserves $13$ human-promoted seed records whose mapping to current rules is not established. Subsequent LLM-assisted proposals and human review filled low-coverage taxonomy cells, including conditional requirements. Detailed source labels remain in the private provenance ledger for audit and rights review; the public release exposes coarse provenance classes and synthetic examples, preventing a direct join from a released rule to a named source repository. The library should therefore be read as empirically grounded benchmark content, not a repository census.


\section{Scoring methods and cascade handling}
\label{sec:app:scoring}

Each rule in an item is scored by one of six methods, selected at authoring time based on what the constraint can be checked against:

\begin{description}
\item[regex.] Surface pattern match on the final-turn message or specific files.
\item[ast.] AST structural match on generated code (e.g., function presence, decorator use, import ordering).
\item[cross-file.] Inter-file consistency check across the generated workspace.
\item[command-output.] Execute a provided compile-or-test script and grade the exit code / stdout.
\item[hybrid.] Deterministic pre-check (narrows candidates) followed by LLM refinement (resolves ambiguity).
\item[LLM-judge.] Rubric-based grading by an LLM judge reading the full run.
\end{description}

For LLM-judge and hybrid methods we use majority voting over three independent evaluations at temperature $0.3$. The frozen coding evaluation uses GPT-5.2 as judge; robustness to this choice is summarized in the main paper's reliability subsection and examined in full in Section~\ref{sec:app:reliability} below.

\paragraph{Severity labels.} Every rule instance carries a severity of \textsc{must}, \textsc{should}, or \textsc{may}, and the released records retain both the unweighted status and the severity-weighted earned/possible totals (weights $3$, $2$, and $1$ respectively). All metrics reported in this paper are unweighted binary rates, so severity affects none of the displayed numbers; it is retained for downstream users who want a weighted view and for the per-item MUST gate used during item authoring.

\paragraph{Denominators and exclusions.} Acc and AP-Acc use pass/fail opportunities only. The frozen panel contains $40{,}104$ method rows: $29{,}176$ pass, $8{,}440$ fail, $2{,}320$ no-opportunity, and $168$ partial. The like-for-like binary recomputation retains the $37{,}616$ pass/fail verdicts; AP-Acc uses the $19{,}449$ eligible verdicts whose final rule label is against-prior. Denominators vary by agent and are retained in the evidence snapshot. These counts apply only to the $12$-model, $60$-item, three-round coding panel; non-coding, E0, and online panels use separate populations and metrics.

\paragraph{All-model common support.} To remove model-specific denominator differences, we additionally restrict to identical $(\mathrm{item},\mathrm{round},\mathrm{rule})$ observations with a clean pass/fail outcome for all $12$ models. This sensitivity retains $2{,}430$ of $3{,}342$ unique observations ($72.7\%$), including $1{,}214$ against-prior observations across $58$ items. We report both views: released-set scores preserve the benchmark panel, while common support enables paired model and Acc--AP-Acc comparisons on an identical denominator.

\paragraph{Cascade-dedup.} A single missing artifact can cause many dependent rules to fail simultaneously. The \_dedup-cascade pass retains one highest-severity fail and converts the remaining dependent outcomes to no-opportunity, preventing one missing artifact from multiplying failures.

\paragraph{Cascade-fairness audit.} After a full batch, audit-cascade-fairness promotes a rule to untestable-design-gap when $\geq 50\%$ of agents miss the artifact needed to test it. Such rules are excluded from denominators. The audit requires at least $5$ tested agents.


\clearpage
\section{Failure decomposition}
\label{sec:app:errors}

The main paper's failure-analysis subsection reports the decomposition by what
the violated rule demands. This section gives the full table, the family
breakdown, and the reason this grouping replaces a free-text taxonomy.

\paragraph{Method.} Every rule in the library carries a logical modality, so each
failure can be assigned to a class from the rule definition alone, with no
inspection of the judge's reason string. Rules that require an action
(\textsc{require}, \textsc{conditional\_require}) or set a floor
(\textsc{limit\_min}) can only be failed by falling short of the demand; rules
that forbid an action (\textsc{forbid}) or set a ceiling (\textsc{limit\_max})
can only be failed by overstepping it; \textsc{prefer} and \textsc{allow} rules
express soft preferences. The decomposition is therefore fully recomputable from
the released verdict records and the released rule definitions, and it involves
no tuned keywords, no per-model adjustment, and no catch-all bucket.

\begin{center}\small
\begin{tabular}{@{}lrrr@{}}
\toprule
Class & Failures & Share & Fail rate \\
\midrule
Shortfall  & $6{,}507$ & 77.1\%  & 23.8\% \\
Overstep   & $1{,}758$ & 20.8\%  & 20.8\% \\
Preference & $175$     & 2.1\%   & 9.4\% \\
\midrule
Total      & $8{,}440$ & 100.0\% & 22.4\% \\
\bottomrule
\end{tabular}
\end{center}

\paragraph{Mass versus propensity.} The share column and the failure-rate column
tell different stories, and only the second is a statement about agent behavior.
Shortfall rules carry $3.7$ times the failure mass of overstep rules, but they
also account for $27{,}306$ of the eligible verdicts against $8{,}443$ for
overstep rules, and the two classes fail at nearly the same rate ($23.8\%$ versus
$20.8\%$). The asymmetry in observed failures is therefore a property of what
operational rule sets ask for---mostly actions---rather than evidence that agents
are markedly more prone to omission than to excess. Soft preferences are the one
class that is genuinely easier, failing at $9.4\%$.

\paragraph{Family-conditional distribution.} Failure mass by family is output
control $27.6\%$, workflow $26.3\%$, code style $15.7\%$, conditional logic
$12.4\%$, tool use $9.5\%$, and quantitative limits $8.6\%$. Output control
combines the largest share of failures with the lowest pass rate; workflow's
share instead reflects its size in the panel, since its pass rate is mid-range. A
pooled accuracy number obscures this family-specific structure.

\paragraph{Relation to the retired keyword taxonomy.} Earlier drafts reported a
twelve-bucket taxonomy assigned by a keyword classifier over judge reason
strings, including a $19.2\%$ unclassified catch-all. We retired it because its
bucket shares could not be regenerated from the released artifacts and because
the reason strings are multilingual free text in which negation cues appear in
nearly every failure, making keyword assignment unreliable. The decomposition
above is reported instead: it is coarser, but every number in it can be
recomputed by a reader from the shipped records.


\section{Additional Experimental Detail}
\label{sec:app:expdetail}

This appendix expands the analyses summarized in the main-text Experiments section. Unless a subsection is explicitly labeled E0, numbers are from the $12$-agent $\times$ $60$-item $\times$ $3$-round coding evaluation ($2{,}160$ scoring records). Coding, non-coding, E0, and online panels are never pooled.

\subsection{Evaluated Model Builds}
\label{sec:app:exp:builds}

Table~\ref{tab:app:builds} gives the model identifier behind each row of the main leaderboard, as logged by the runner at collection time rather than reconstructed afterwards. Where a provider published a dated snapshot, the date is part of the identifier. The three Anthropic builds were served with their extended $1$M-context variant. Two entries were preview releases when the panel was collected and have since been promoted, so the current catalogue names differ from what was run---\texttt{qwen/qwen3.6-max-preview} is now \texttt{qwen/qwen3.6-max}, and \texttt{tencent/hy3-preview:free} is now \texttt{tencent/hy3}; we record what was run. Seed-2.0-Pro was reached through a provider-hosted deployment rather than a shared catalogue route, so the table gives its model identifier rather than a routing slug. The judge is \texttt{openai/gpt-5.2} throughout, at temperature $0.3$ with three-vote majority.

\begin{table}[htbp]
\centering
\small
\caption{Model identifier behind each leaderboard row, as logged at collection time.}
\label{tab:app:builds}
\begin{tabular}{@{}ll@{}}
\toprule
\textbf{Displayed name} & \textbf{Model identifier} \\
\midrule
Claude-Opus-4.7   & \texttt{anthropic/claude-4.7-opus-20260416} \\
Claude-Sonnet-4.6 & \texttt{anthropic/claude-4.6-sonnet-20260217} \\
Claude-Haiku-4.5  & \texttt{claude-haiku-4-5-20251001} \\
GPT-5.5           & \texttt{openai/gpt-5.5} \\
Gemini-3.1-Pro    & \texttt{google/gemini-3.1-pro-preview} \\
Qwen-3.6-Max      & \texttt{qwen/qwen3.6-max-preview} \\
Hy3               & \texttt{tencent/hy3-preview:free} \\
Kimi-K2.6         & \texttt{moonshotai/kimi-k2.6} \\
MiniMax-M2.7      & \texttt{minimax/minimax-m2.7} \\
GLM-5.1           & \texttt{z-ai/glm-5.1} \\
StepFun-3.5       & \texttt{stepfun/step-3.5-flash} \\
Seed-2.0-Pro      & \texttt{seed-2.0-pro} \\
\bottomrule
\end{tabular}
\end{table}

\subsection{Per-Surface and Per-Family Accuracy}
\label{sec:app:exp:cells}

Table~\ref{tab:app:channel-csr} reports pooled accuracy for each configurable surface, and Table~\ref{tab:app:family-csr} reports pooled accuracy for each constraint family. Both are recomputed from the released panel over the same $37{,}616$ eligible verdicts as the main table. The user-instruction (UI) surface carries only against-prior placements in this panel ($1{,}476$ of $1{,}476$ eligible verdicts), so its rate is not comparable with the other surfaces, whose placements are mixed; it is reported for completeness rather than as a surface-difficulty estimate.

\begin{table}[!ht]
\centering
\small
\caption{Pooled accuracy by instruction surface, with the number of eligible verdicts. PF is the project file (e.g.\ \texttt{CLAUDE.md}); SD is the skill description; TD the tool description; SP the system prompt; UI the user instruction.}
\label{tab:app:channel-csr}
\begin{tabular}{@{}lrr@{}}
\toprule
Surface & Pooled Acc & $N$ \\
\midrule
TD (tool description)    & $83.1\%$ & $3{,}934$ \\
PF (project file)        & $79.1\%$ & $11{,}756$ \\
SD (skill description)   & $78.6\%$ & $11{,}072$ \\
SP (system prompt)       & $73.6\%$ & $6{,}589$ \\
UI (user instruction)\textsuperscript{$\dagger$} & $54.5\%$ & $1{,}476$ \\
\bottomrule
\end{tabular}

\smallskip
{\footnotesize \textsuperscript{$\dagger$}All UI placements in this panel are against-prior, unlike every other surface.}
\end{table}

\begin{table}[!ht]
\centering
\small
\caption{Pooled accuracy by constraint family, with the number of eligible verdicts and the per-build range. Output control is lowest overall and for $11$ of the $12$ builds; quantitative limits are highest.}
\label{tab:app:family-csr}
\begin{tabular}{@{}lrrl@{}}
\toprule
Family & Mean Acc & $N$ & Per-build range \\
\midrule
QT (quantitative)     & $82.6\%$ & $4{,}168$ & $74.9$--$92.4$ \\
TU (tool use)         & $81.6\%$ & $4{,}352$ & $77.7$--$90.6$ \\
CL (conditional)      & $80.9\%$ & $5{,}489$ & $78.1$--$88.2$ \\
CS (code style)       & $79.2\%$ & $6{,}342$ & $68.8$--$93.5$ \\
WF (workflow)         & $76.0\%$ & $9{,}265$ & $71.0$--$84.1$ \\
OC (output control)   & $70.9\%$ & $8{,}000$ & $63.9$--$78.9$ \\
\bottomrule
\end{tabular}
\end{table}

\FloatBarrier
\subsection{Cross-Agent Agreement and Clusters}
\label{sec:app:exp:cluster}

Agents largely agree on which constraints are hard, though the strength of that agreement depends on the comparison used. Correlating each build's per-rule pass-rate vector with the cohort mean over the $242$ rules every build attempted gives $0.57$--$0.89$ (mean $0.80$); the stricter build-to-build pairwise correlation gives $0.32$--$0.83$ (mean $0.62$). The leaderboard therefore reflects a largely shared difficulty structure with real per-build idiosyncrasy on individual rules.

Under the two-cluster grouping recorded in the release ($5$ builds versus $7$), mean accuracy differs by $+5.9$ points ($81.2\%$ versus $75.3\%$) and mean AP-Acc by $+5.5$ points ($75.1\%$ versus $69.7\%$). Prior control therefore does not explain the cluster difference: the gap is almost unchanged when the comparison is restricted to against-prior rules. Pairwise cosine similarity between per-build modality$\times$surface profile vectors stays in $[0.987,0.999]$ over the $16$ cells with at least $20$ eligible verdicts, so the clusters differ in level rather than in the shape of their compliance profile, and we treat them as a weak behavioral gradient rather than discrete agent types.

\subsection{Separate E0 Surface-Precedence Pilot}
\label{sec:app:exp:e0}

The main surface-rank figure comes from E0, a distinct pilot collected on 2026-04-25 using nine older model builds, $916$ recorded runs, four synthetic conflict pairs, and deterministic-only scoring. The nine builds are Claude Opus 4.6, Claude Sonnet 4.6, GPT 5.4, Gemini 3.1 Pro, DeepSeek V3.2, Kimi K2.5, MiniMax M2.7, Qwen 3.6 Plus, and Seed 2 Pro; four of them do not appear in the main coding panel, which is one reason E0 is not pooled with it. E0 predates the current surface naming; its two legacy channel labels map onto the canonical set as CM~$\to$~project file (PF) and SK~$\to$~skill description (SD), and we report E0 in canonical terms throughout. Ordinal ranks are recomputed from each model's head-to-head win counts and averaged equally across models despite unequal focused rerun counts: SP/PF/UI $=2.22$, TD $=3.78$, and SD $=4.56$. These ranks describe precedence under the synthetic conflicts, not coding accuracy, and are not pooled with the main $12$-model panel. The two views therefore order the skill description differently and are not in conflict: E0 asks which surface wins when two surfaces demand opposite things, and places SD last; the main panel asks how often a rule is followed when it is the only instruction, where SD reaches $78.6\%$ above SP at $73.6\%$. A surface can be easy to comply with in isolation and still lose a direct conflict, so the E0 ranking is a statement about precedence and the panel rates are statements about difficulty.

\paragraph{Identification and uncertainty.}
E0 uses counterbalanced assignment rather than paired outcomes: each AB and BA run is a separate fresh session, with the mutually exclusive members of one rule pair assigned to opposite surfaces for the same fixed task. The design contains $458$ assigned runs per direction; after $27$ errors, $445$ AB and $444$ BA runs are decisive. On these $889$ outcomes, a pooled Bradley--Terry fit gives sum-to-zero log-strengths SP $=+0.57$, PF $=+0.65$, UI $=+0.60$, TD $=-0.53$, and SD $=-1.29$. Because model and conflict pair are crossed factors, the primary uncertainty analysis independently resamples the nine models and four pairs ($10{,}000$ fixed-seed replicates), rather than treating $36$ model--pair cells as independent clusters. The complete SP/PF/UI $>$ TD $>$ SD ordering appears in $9{,}652/10{,}000$ crossed-bootstrap resamples (top-group-above-TD: $98.81\%$; TD-above-SD: $97.71\%$); these are Monte Carlo diagnostics conditional on the resampling scheme, not $p$-values. The three leading surfaces are not statistically distinguished at the available precision.

The pooled ordering survives both direction-specific fits ($445/444$ decisive rows), all four leave-one-pair-out fits, all nine leave-one-model-out fits, and equal weighting of the $694$ model$\times$pair$\times$direction$\times$surface-matchup cells (repeat range $1$--$5$). It also survives four deterministic assignments of every error---channel A wins, channel B wins, the higher expected group wins, or the lower expected group wins. Errors are concentrated in DeepSeek V3.2 ($14$) and Seed 2 Pro ($10$), so this explicit bound is preferable to assuming outcome-independent deletion. Heterogeneity remains material: all four pair-only fits, but only six of nine model-only fits, reproduce the exact ordering. We therefore interpret E0 as a robust pooled tendency within four synthetic style conflicts and nine older builds, not a build-invariant hierarchy or evidence for the main panel's descriptive surface strata. AB was executed before BA in the retained collection; the counterbalanced assignment addresses rule identity but the fixed temporal order cannot be corrected retrospectively.

The pooled ordering survives both direction-specific fits ($445/444$ decisive rows), all four leave-one-pair-out fits, all nine leave-one-model-out fits, and equal weighting of the $694$ model$\times$pair$\times$direction$\times$surface-matchup cells (repeat range $1$--$5$). It also survives four deterministic assignments of every error---channel A wins, channel B wins, higher expected tier wins, or lower expected tier wins. Errors are concentrated in DeepSeek V3.2 ($14$) and Seed 2 Pro ($10$), so this explicit bound is preferable to assuming outcome-independent deletion. Heterogeneity remains material: all four pair-only fits, but only six of nine model-only fits, reproduce the exact tier ordering. We therefore interpret E0 as a robust pooled tendency within four synthetic style conflicts and nine older builds, not a build-invariant hierarchy or evidence for the main panel's descriptive surface strata. AB was executed before BA in the retained collection; the counterbalanced assignment addresses rule identity but the fixed temporal order cannot be corrected retrospectively.

\subsection{Surface $\times$ Family Interaction}
\label{sec:app:exp:interaction}

The main coding panel does not provide a paired surface intervention: constraints are assigned to admissible surfaces rather than counterbalanced across all placements, and surface wording and family composition can co-vary. We therefore treat its surface-by-family cells as descriptive strata rather than a variance decomposition or controlled surface effect. A historical held-out analysis reports $648$ additional conflict runs, Kendall's $\tau=0.91$, and an $89\%$ stronger-group win rate, but its row-level artifact is not retained. We treat those figures as descriptive historical evidence only, not as a robustness result.


\section{Full reliability analysis}
\label{sec:app:reliability}

\begin{figure}[!ht]
\centering
\includegraphics[width=0.95\linewidth]{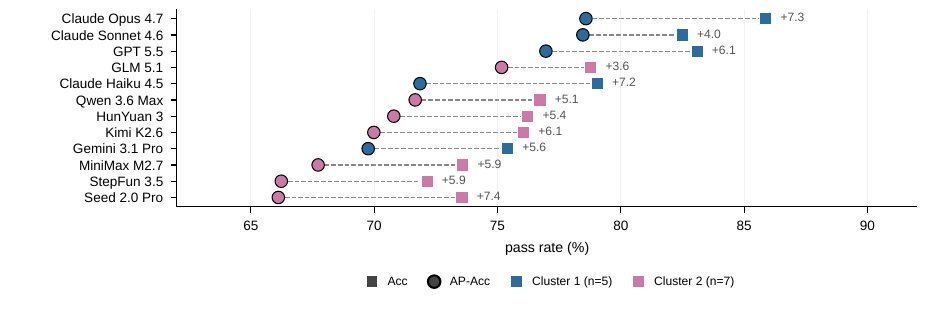}
\caption{Per-agent Accuracy (square) and AP-Acc (circle), sorted by AP-Acc, both recomputed from the released panel under the same binary definition. The dashed span is the Acc--AP-Acc prior-alignment gap, positive for all $12$ agents. Marker fill encodes the two behavioral clusters described under experimental detail.}
\label{fig:inflation}
\end{figure}

The main paper's reliability subsection summarizes the reliability of the reported results. This appendix provides the full stratified numbers, ablations, and method details. Figure~\ref{fig:inflation} shows the reported per-model Accuracy and AP-Acc and the descriptive gap between them.

\subsection{Judge calibration and judge-swap sensitivity}

The frozen coding panel uses a 3-vote protocol for rubric and hybrid methods, with the majority outcome as the consensus verdict. Two retained calibration studies answer different questions. First, on $65$ common non-error samples, Claude--GPT four-class inter-LLM agreement is $\kappa=0.4717$, whereas auto-vs-Claude is $0.0332$ and auto-vs-GPT is $-0.0491$; these are inter-LLM comparisons, not human-reference results, and expose sensitivity around no-opportunity boundaries. Second, a historical human-reference audit of a related five-vote auto-scoring configuration records a three-class confusion matrix over $n=919$ rows, with $69.0\%$ observed agreement and $\kappa=0.515$. This provides moderate human-reference calibration for the judge family, not direct validation of the exact frozen three-vote panel. The matrix arithmetic and per-case disagreement reasons are retained, but the raw joined labels, the sampling frame and probabilities, rater assignments, and the independent adjudication record are not, so it is not a fully auditable human calibration study and we report it as qualified evidence only.

\paragraph{Judge-swap ablation (Claude Opus 4.7 in place of GPT 5.2).}
We re-judge a stratified 200-verdict subset (balanced across family $\times$ modality $\times$ prior cells) with Claude Opus 4.7 as the judge, using the same 3-vote protocol. Both judges produce a clean pass/fail/partial label for 116 rows; 84 rows fall outside the paired comparison because at least one side is no-opportunity, an error, or lacks evidence. On the paired subset, raw agreement is $62.1\%$ and Cohen's $\kappa = 0.163$; per-agent pass-rate deltas range from $-40$\,pp (Seed) to $+33$\,pp (Kimi, MiniMax). This sensitivity motivates our emphasis on cross-model patterns and common-support uncertainty rather than narrow leaderboard margins. The retained release includes the aggregate judge-swap summary, but not row-level swap verdicts, so an alternate-judge prior-alignment gap is unavailable.

\subsection{Test-retest (ICC) detail}

\begin{table}[!ht]
\centering
\small
\caption{Test-retest stability across the three rounds, by stratum.}
\label{tab:app:icc}
\begin{tabular}{@{}lr@{}}
\toprule
Stratum & ICC(1,1) \\
\midrule
Agent-level (12 agents, pooled cells)            & $0.725$ \\
Cell-level (agent $\times$ item)                 & $0.599$ \\
Mean per-cell Acc range (across rounds)          & $15.6$\,pp \\
\bottomrule
\end{tabular}
\end{table}

Round-to-round dispersion is substantial at the cell level: the mean per-cell Acc
range across the three rounds is $15.6$ points, which exceeds the $13.7$-point
spread of the whole leaderboard. Single-round comparisons between adjacent models
are therefore uninformative, and all reported numbers pool three rounds.

\subsection{Scoring-method composition and ranking robustness}

Method shares are computed over the $37{,}616$ eligible pass/fail verdicts of the
released panel. Kendall $\tau$ compares each method-specific per-agent ranking
with the overall ranking; $N$/agent is the mean number of eligible verdicts per
agent. The judge and the deterministic--judge hybrid together cover $86.7\%$ of
eligible verdicts, so the judge-swap sensitivity reported above applies to most
of the measurement.

\begin{table}[!ht]
\centering
\small
\setlength{\tabcolsep}{4pt}
\caption{Per-method agreement with the overall ranking (Kendall's $\tau$), mean eligible verdicts per agent, each method's share of the $37{,}616$ eligible verdicts, and its pass rate. The judge and the deterministic--judge hybrid together cover $86.7\%$ of eligible verdicts.}
\label{tab:app:method-robust}
\begin{tabular}{@{}lcrrr@{}}
\toprule
Method & $\tau$ vs.\ overall & $N$/agent & Share & Pass rate \\
\midrule
LLM-judge      & $+0.91$ & $2{,}147$ & $68.5\%$ & $72.6\%$ \\
hybrid         & $+0.72$ & $570$     & $18.2\%$ & $90.8\%$ \\
regex          & $+0.79$ & $275$     & $\phantom{0}8.8\%$  & $86.1\%$ \\
command-output & $+0.39$ & $\phantom{0}84$      & $\phantom{0}2.7\%$  & $84.1\%$ \\
ast            & $+0.23$ & $\phantom{0}31$      & $\phantom{0}1.0\%$  & $89.3\%$ \\
cross-file     & $+0.00$ & $\phantom{0}28$      & $\phantom{0}0.9\%$  & $76.3\%$ \\
\bottomrule
\end{tabular}
\end{table}

\subsection{Prior-label lineage and sensitivity}

The current governed distribution is $115$ align-prior, $282$ against-prior, and $245$ neutral. The recoverable $5/9$ zero-injection consensus covers $287$ rules ($106/170/11$); a separate historical report describes a $280$-rule binary determined subset ($114/166$), which is not interchangeable with either the consensus count or the final $282$ against-prior labels. Lineage matches the zero-injection consensus for $275$ rules, matches a recoverable pre-existing value for $331$, and remains unknown for $36$; $12$ final labels differ from recoverable consensus without a preserved override reason.

\paragraph{Probe cohort overlap with the evaluated panel.}
Because the against-prior set is defined partly from observed model behavior, we state how much that behavior overlaps the models being scored. The zero-injection votes come from a nine-build probe cohort recorded in the frozen lineage snapshot, which ships with the release. Seven of the twelve evaluated models---Claude~Haiku~4.5, GLM-5.1, GPT-5.5, Hy3, Kimi~K2.6, Qwen~3.6~Max, and StepFun-3.5---are absent from that cohort and therefore contributed no prior labels. The against-prior gap is positive for all seven, averaging $+5.63$ points (range $+3.62$ to $+7.19$), against $+6.06$ points for the five that share an identifier with a probe build. The $0.43$-point difference between the two groups is small relative to the effect itself and, on twelve models, is not resolved in either direction; we report it as a bound rather than as evidence of independence. Two further limits apply. Because only four probe builds sit outside the evaluated panel's identifier set, a $5/9$ consensus necessarily includes at least one overlapping build, so no zero-injection label is independent of the scored cohort. And absence from the probe cohort is not vendor independence: GPT-5.5, Kimi~K2.6, and Qwen~3.6~Max are successor builds of probe models, leaving GLM-5.1, Hy3, and StepFun-3.5 as the only evaluated models with no same-vendor probe build; their mean gap is $+4.99$ points. What does bound the exposure is the label provenance itself: of the $19{,}449$ against-prior eligible verdicts, $8{,}574$ ($44.1\%$) carry a label sourced from the zero-injection consensus, against $9{,}051$ ($46.5\%$) from prior curation and $1{,}824$ ($9.4\%$) of unknown lineage. Fewer than half of the AP-Acc denominator can therefore be affected by the overlap at all.

Under the paper-like binary equations, the frozen full panel contains $37{,}616$ eligible pass/fail verdicts, of which $19{,}449$ are against-prior; no-opportunity and other statuses are excluded. The resulting agent-macro mean Acc--AP-Acc gap is $+5.8076$ points and is positive for all $12$ agents. The deterministic-only subset contains $5{,}013$ eligible verdicts ($977$ against-prior) and gives $+13.0933$ points, also positive for all agents; larger historical estimates are not retained.

The available threshold sweep relabels the recoverable zero-injection consensus subset. Under the same binary definition, the mean gap is $+11.8161$ points at $4/9$ and $5/9$, $+10.6022$ at $6/9$, and $+10.2438$ at $7/9$; the $4/9$--$6/9$ change is $1.2139$ points. The direction is therefore stable across the tested thresholds.

\subsection{Common-support uncertainty}

Common support requires a clean pass/fail outcome from every model for the same $(\mathrm{item},\mathrm{round},\mathrm{rule})$ key. It retains $2{,}430/3{,}342$ unique observations ($72.7\%$), including $1{,}214$ against-prior observations over $58$ items. We use a deterministic $2{,}000$-resample item-clustered percentile bootstrap (seed $20260723$), retaining all common-support rounds and rule rows within each sampled item. The paired Acc--AP-Acc interval is positive for every model; lower bounds range from $+1.06$ to $+4.83$ points. By contrast, all adjacent common-support Acc intervals include zero, yielding one conservative adjacent tie group containing all $12$ models. Thus the prior-alignment contrast is resolved model by model, whereas neighboring point ranks are not.


\section{Failure gallery}
\label{sec:app:gallery}

Five failures from the frozen scoring panel illustrate the benchmark's main findings. To preserve source abstraction, the vignettes use paper-local labels and behavioral paraphrases rather than stable library identifiers or verbatim constraint text.

\begingroup
\footnotesize\sloppy
\begin{description}
\item[Example A: skill-surface placement miss.] Two agents created the requested skill artifact but placed it outside the required project convention. The same rule was followed more often when delivered through a project file ($100\%$ of $144$ eligible verdicts) than through the skill description ($67.0\%$ of $103$). These are distinct rule instances in distinct items rather than a counterbalanced pair, so the contrast is descriptive.

\item[Example B: tool-surface workflow miss.] An agent produced a free-form change summary rather than the required structured form when the workflow rule appeared in a tool description. The rule is placed only on the tool description in this panel, where it is followed in $91.9\%$ of its $172$ eligible verdicts, so this is an instance failure on an otherwise well-followed rule rather than evidence about the surface.

\item[Example C: against-prior output failure.] Two agents generated an additional documentation artifact despite an explicit prohibition. The task naturally invited explanatory output, so compliance required overriding a common default rather than merely completing the requested implementation.

\item[Example D: output-control mirroring.] An agent copied language and formatting cues from the surrounding fixture into its final response, violating a response-language constraint. This reflects the surface-mirroring pattern common in the output-control family.

\item[Example E: budget overshoot.] Two agents produced artifacts substantially beyond a stated size cap. The result illustrates failure on a quantitative constraint when unconstrained generation favors longer outputs.
\end{description}
\endgroup


\section{Exploratory Non-Coding Extension}
\label{sec:app:noncoding}

\paragraph{Scope and protocol.}
We evaluated 40 non-coding cases, eight in each of five domains: customer support, legal/compliance, marketing content, financial analysis, and research/academic writing. The panel comprises 12 models, each assigned three generator runs per case (1,440 attempted trajectories). These tasks primarily produce natural-language artifacts rather than repository patches; their language, tools, interaction horizons, and rule ontology therefore differ from the coding panel. The model builds are those recorded by the frozen Stage-5 evaluation, not necessarily the identically named builds in later panels; two rows use the Stage-5 display names Hunyuan~3 and StepFun~Flash, which correspond to the coding leaderboard's Hy3 and StepFun-3.5.

\paragraph{Metric.}
For accepted run $r$ of case $c$ and model $m$, let $E_{cmr}$ be the eligible rule verdicts after the evaluation's exclusions and let $p_{cmr}=|E_{cmr}|^{-1}\sum_{j\in E_{cmr}}\mathbf{1}[j=\textsc{pass}]$. We first average $p_{cmr}$ over the valid runs for each case--model pair, then average the 40 case means equally:
\begin{equation}
  \mathrm{NC\mbox{-}Macro}_m = \frac{1}{40}\sum_{c=1}^{40}
  \frac{1}{|R_{cm}|}\sum_{r\in R_{cm}} p_{cmr}.
\end{equation}
This case-macro construction prevents cases with more rubric checks from dominating. Empty-shell trajectories are excluded; a non-empty trajectory with at least 25\% judge errors uses its clean eligible-rule rate, following the frozen evaluation protocol. Confidence intervals in Table~\ref{tab:noncoding-overall} are percentile intervals from 20,000 fixed-seed ($20260714$), domain-stratified bootstrap replicates that resample cases within each domain and retain each sampled case's observed run aggregate. Cases, rather than runs or individual rule verdicts, are the resampling units.

\begin{table}[!ht]
\centering
\small
\setlength{\tabcolsep}{3.2pt}
\caption{Exploratory non-coding results. NC-Macro and its 95\% case-bootstrap interval are percentages. $\sigma_r$ is the mean within-case run standard deviation; $\sigma_c$ is the cross-case standard deviation. Valid is out of 120 attempted trajectories per model.}
\label{tab:noncoding-overall}
\begin{tabular}{@{}lrrrr@{}}
\toprule
Model & NC-Macro [95\% CI] & $\sigma_r$ & $\sigma_c$ & Valid \\
\midrule
GPT 5.5             & 84.8 [83.0, 86.6] & 3.0 & 6.6 & 120 \\
GLM 5.1             & 78.7 [76.6, 80.7] & 4.5 & 8.2 & 120 \\
Claude Opus 4.7     & 78.4 [75.4, 81.2] & 4.2 & 9.8 & 116 \\
Gemini 3.1 Pro      & 76.9 [74.5, 79.1] & 6.1 & 8.9 & 120 \\
Claude Sonnet 4.6   & 76.1 [73.5, 78.6] & 5.3 & 8.9 & 119 \\
Hunyuan 3           & 73.3 [70.4, 76.4] & 6.5 & 10.4 & 120 \\
Qwen 3.6 Max        & 72.4 [69.7, 74.9] & 6.3 & 9.0 & 119 \\
Claude Haiku 4.5    & 71.3 [69.0, 73.5] & 6.0 & 8.5 & 120 \\
Seed 2 Pro          & 69.6 [66.6, 72.7] & 4.5 & 11.7 & 120 \\
Kimi K2.6           & 68.6 [65.3, 71.9] & 7.6 & 11.4 & 114 \\
MiniMax M2.7        & 68.5 [65.6, 71.3] & 6.2 & 10.3 & 120 \\
StepFun Flash       & 65.8 [63.0, 68.6] & 6.6 & 10.8 & 120 \\
\bottomrule
\end{tabular}
\end{table}

\begin{table}[!ht]
\centering
\small
\setlength{\tabcolsep}{2.4pt}
\caption{Case-macro pass rate (\%) by non-coding domain; each cell averages eight case means. Model order follows overall NC-Macro, not the coding leaderboard.}
\label{tab:noncoding-domains}
\begin{tabular}{@{}lrrrrr@{}}
\toprule
Model
  & \multicolumn{1}{c}{\shortstack{Customer\\support}}
  & \multicolumn{1}{c}{\shortstack{Legal/\\compliance}}
  & \multicolumn{1}{c}{Marketing}
  & \multicolumn{1}{c}{\shortstack{Financial\\analysis}}
  & \multicolumn{1}{c}{\shortstack{Research/\\academic}} \\
\midrule
GPT 5.5             & 80.5 & 84.8 & 84.9 & 90.2 & 83.7 \\
GLM 5.1             & 73.7 & 82.2 & 77.7 & 85.8 & 73.8 \\
Claude Opus 4.7     & 76.8 & 80.2 & 75.6 & 81.9 & 77.3 \\
Gemini 3.1 Pro      & 72.4 & 77.7 & 77.8 & 85.1 & 71.3 \\
Claude Sonnet 4.6   & 72.5 & 78.1 & 73.2 & 81.4 & 75.3 \\
Hunyuan 3           & 72.2 & 70.7 & 71.7 & 80.9 & 71.1 \\
Qwen 3.6 Max        & 71.2 & 72.3 & 74.5 & 75.9 & 67.9 \\
Claude Haiku 4.5    & 67.0 & 68.1 & 72.4 & 78.8 & 70.1 \\
Seed 2 Pro          & 69.4 & 68.5 & 73.6 & 77.5 & 59.3 \\
Kimi K2.6           & 71.8 & 71.7 & 66.2 & 71.4 & 62.1 \\
MiniMax M2.7        & 65.7 & 66.9 & 67.5 & 77.5 & 64.8 \\
StepFun Flash       & 64.3 & 65.7 & 64.6 & 75.8 & 58.4 \\
\bottomrule
\end{tabular}
\end{table}

\paragraph{Coverage and variability.}
Of 1,440 attempted trajectories, 1,428 (99.2\%) were valid. The 12 invalid trajectories were empty long-context network shells and were excluded rather than scored. Eleven case--model cells consequently have incomplete replication: ten retain two valid runs and one retains a single valid run. Their means use the available accepted runs; $\sigma_r$ uses the observed sample size and is undefined for the single-run cell. Table~\ref{tab:noncoding-overall} reports both run-wise and cross-case variation because the latter is generally larger and the 40 cases, not the 1,428 trajectories, define the external-validity sample. Per-domain case-macro pass rates appear in Table~\ref{tab:noncoding-domains}.

\paragraph{Non-comparability and provenance.}
NC-Macro is not AP-Acc: it uses a different task population, ontology, scoring/exclusion protocol, and model-build snapshot. We therefore neither pool the 40 cases with the 60 coding items nor use this table to revise the paper's coding leaderboard. This historical analysis is also distinct from the July 2026 online archive, which contains new-build, single-trial operational runs and is not an additional round of either panel. The values above come from a frozen 12-model Stage-5 aggregate snapshot preserved in the release provenance. Complete raw trajectories no longer survive in the current repository, so the snapshot supports the stated aggregate-level claims while trajectory-level regeneration remains outside scope. This provenance boundary, together with protocol and build differences, makes the analysis exploratory rather than confirmatory.


\section*{Datasheet}
\label{sec:datasheet}

Following \citet{gebru2021datasheets}.

\paragraph{Motivation.} The benchmark was created to evaluate instruction following in coding agents across multiple instruction-placement surfaces (system prompt, tool description, skill description, project file, user message, plus a harness default)---a gap left by existing IF benchmarks that only test user-message prompts.

\paragraph{Composition.} The $80$-candidate disposition retains $60$ items, deprecates $15$, and excludes $5$ after quality review. Several per-item promotion and review artifacts are not recoverable. The diagnostic sampler is separate from and does not define the final assembled items. The release contains three explicitly separate evaluation populations. The frozen coding core comprises: (i) a library of 642 atomic constraints authored in YAML and tagged along 8 axes (family, modality, prior, observability, verifiability, universality, surface fit, and surface variants); (ii) 60 assembled coding items (quality-audited from an 80-item working set), each combining a scenario fixture, an injected pack of 25--35 rules of which 10--27 are scorable given the opportunities the item creates, a surface assignment, a multi-turn task specification, and a ground-truth scoring script; across the panel 302 distinct library rules are instantiated and 256 receive at least one verdict, and professional-writing rules are exercised only in the non-coding panel; and (iii) 2{,}160 scoring records ($12$ agents $\times 60$ coding items $\times 3$ rounds) with per-rule pass/fail/no-opportunity status and judge reasons. A frozen exploratory non-coding panel separately contains 40 cases across five domains (eight cases each), with 1{,}440 attempted trajectories ($12$ models $\times 40$ cases $\times 3$ generator runs) and 1{,}428 valid trajectories. Its case-macro rule pass rate averages valid runs within each case and then weights the 40 case means equally; it is not pooled with coding AP-Acc. A July 2026 online operational audit is a third, new-build single-trial panel covering 60 coding items and 40 general cases; outages and API-quality flags are retained, and its metrics are not comparable to either frozen panel. Constraints include natural-language text; items include source code fixtures and task specifications.

\paragraph{Collection process.} A hash-verified manual survey of publicly accessible GitHub project-instruction and contributor documents seeded the constraint library. Surveyed material included \texttt{CLAUDE.md}, \texttt{AGENTS.md}, \texttt{CONTRIBUTING.md}, skill descriptions, and tool schemas across multiple software stacks. Requirements were atomized and normalized into project-authored benchmark constraints rather than copied as repository files. LLM-assisted proposals and author review subsequently filled taxonomy gaps. A later commit preserves 13 human-promoted historical seed records in Git history, although their mapping to current rules is not recoverable, so the retained labels do not support a rule-level transformation ledger. Detailed source labels are retained privately for audit and rights review; the public artifact uses coarse provenance classes and synthetic paper examples rather than repository-shaped labels or source-searchable quotations. Coding workspaces were assembled for realistic technology stacks and are project-authored, owner-attested on 2026-07-27 after a mechanical provenance audit of all 785 fixture files found no copied third-party material; earlier drafts held them back pending that provenance and redistribution rights are cleared. No crowdworkers were used.

\paragraph{Preprocessing / cleaning / labelling.} The retained evidence does not support a comprehensive construction-time deduplication or PII-removal claim. Current prior labels comprise 115 \texttt{align\_prior}, 282 \texttt{against\_prior}, and 245 \texttt{neutral} rules. A reproducible zero-injection $5/9$ consensus covers a distinct 287-rule subset; a historical report's 280-rule determined subset is reported-only and lacks a retained row-level manifest. Where label lineage or override rationale is not recoverable, the frozen lineage snapshot records it as unknown rather than inferring it.

\paragraph{Uses.} Intended for benchmarking instruction-following performance of coding agents and for validating surface-aware and prior-aware evaluation methodologies. Not intended as a certification instrument, agent capability certification, or hiring/procurement decision tool.

\paragraph{Data availability.} A planned public release will contain the $642$ project-authored constraints, active item specifications, surface assignments, ground-truth scoring scripts, sanitized derived verdicts for the frozen $2{,}160$-record coding panel, analysis code and snapshots, Croissant metadata, and provenance documentation. The executable coding workspaces are included following the 2026-07-27 owner attestation; raw provider traces remain excluded until their privacy review is complete, which affects end-to-end trajectory replay but not inspection of the constraint library, item design, scoring logic, or published aggregate analyses.

\paragraph{Distribution.} The release does not grant rights to or publicly redistribute raw third-party repository snapshots, provider outputs, request traces, or unreviewed third-party material. Project-authored software is intended for distribution under Apache-2.0 and project-authored benchmark text/data plus sanitized derived results under CC-BY-4.0; the file-level rights ledger and exporter allowlists govern actual inclusion.

\paragraph{Maintenance.} Maintained by the authors; after public release, issues and contributions will be handled through the public repository. The benchmark may be updated (new constraints, items, or agent evaluations); versioned via manifest.yaml.

\paragraph{Known biases.} The constraint library reflects an English-language, software-engineering-centered manual survey plus targeted expansion; free-form source labels do not establish a repository sampling frame. The 12-agent evaluation sample consists of generally available commercial and public API models at time of writing; it does not include specialized coding-only models or fine-tuned variants. Coding-item selection used observed pilot/full-panel discriminativeness and difficulty diagnostics, which may favor separation among the observed models (selection optimism). Any training-provenance correspondence in model behavior should be interpreted descriptively rather than causally.

\paragraph{Legal and ethical considerations.} Human reviewers assessed model outputs and automated verdicts during quality control; the retained historical comparison is reported as qualified calibration evidence, and no rater identities or personal data are released. The evaluation did not collect behavioral or demographic data from human subjects. Public exports must pass fail-closed privacy and rights checks and exclude credentials, private endpoints, request identifiers, raw provider responses, hidden reasoning, unreviewed personal data, and material without established redistribution rights. Public availability of a source does not by itself establish redistribution permission.

\end{document}